\documentclass[sigconf, screen]{acmart}
\AtBeginDocument{%
  }

\setcopyright{acmlicensed}
\copyrightyear{2024}
\acmYear{2024}
\acmDOI{XXXXXXX.XXXXXXX}

\acmConference[Arxiv
Preprint]{}{August 22,
  2024}{}
\usepackage{microtype}
\usepackage{hyperref}
\usepackage{url}
\usepackage{booktabs}
\usepackage{makecell}
\usepackage{tabularx}
\usepackage{caption}
\setcellgapes{5pt}
\definecolor{darkblue}{rgb}{0, 0, 0.5}
\hypersetup{colorlinks=true, citecolor=darkblue, linkcolor=darkblue, urlcolor=darkblue}
\usepackage{colortbl}
\usepackage{multicol}

\begin{document}

%%
%% The "title" command has an optional parameter,
%% allowing the author to define a "short title" to be used in page headers.
\title{Investigating LLM Applications in E-Commerce}

%%
%% The "author" command and its associated commands are used to define
%% the authors and their affiliations.
%% Of note is the shared affiliation of the first two authors, and the
%% "authornote" and "authornotemark" commands
%% used to denote shared contribution to the research.
\author{Chester Palen-Michel}
\authornote{Each author contributed equally to this research.}
\email{cpalenmichel@brandeis.edu}

\affiliation{%
  \institution{Mitchom School of Computer Science  Brandeis University}
  \city{Waltham}
  \state{Massachusetts}
  \country{USA}
}

\author{Ruixiang Wang}
\authornotemark[1]
\email{ruixwang@ebay.com}

\affiliation{%
  \institution{eBay}
  \city{San Jose}
  \state{California}
  \country{USA}}

\author{Yipeng Zhang}
\authornotemark[1]
\email{yipezhang@ebay.com}

\affiliation{%
  \institution{eBay}
  \city{San Jose}
  \state{California}
  \country{USA}}

\author{David Yu}
\email{hongjyu@ebay.com}

\affiliation{%
  \institution{eBay}
  \city{San Jose}
  \state{California}
  \country{USA}}

\author{Canran Xu}
\email{canxu@ebay.com}

\affiliation{%
  \institution{eBay}
  \city{San Jose}
  \state{California}
  \country{USA}}

\author{Zhe Wu}
\email{zwu1@ebay.com}

\affiliation{%
  \institution{eBay}
  \city{San Jose}
  \state{California}
  \country{USA}}

%%
%% By default, the full list of authors will be used in the page
%% headers. Often, this list is too long, and will overlap
%% other information printed in the page headers. This command allows
%% the author to define a more concise list
%% of authors' names for this purpose.
\renewcommand{\shortauthors}{Palen-Michel et al.}

%%
%% The abstract is a short summary of the work to be presented in the
%% article.
\begin{abstract}
  The emergence of Large Language Models (LLMs) has revolutionized natural language processing in various applications especially in e-commerce. 
One crucial step before the application of such LLMs in these fields is to understand and compare the performance in different use cases in such tasks. 
This paper explored the efficacy of LLMs in the e-commerce domain, focusing on instruction-tuning an open source LLM model with public e-commerce datasets of varying sizes and comparing the performance with the conventional models prevalent in  industrial applications. 
We conducted a comprehensive comparison between LLMs and traditional pre-trained language models across specific tasks intrinsic to the e-commerce domain, namely classification, generation, summarization, and named entity recognition (NER). Furthermore, we examined the effectiveness of the current niche industrial application of very large LLM, using in-context learning, in e-commerce specific tasks. Our findings indicate that few-shot inference with very large LLMs often does not outperform fine-tuning smaller pre-trained models, underscoring the importance of task-specific model optimization.Additionally, we investigated different training methodologies such as single-task training, mixed-task training, and LoRA merging both within domain/tasks and between different tasks. 
Through rigorous experimentation and analysis, this paper offers valuable insights into the potential effectiveness of LLMs to advance natural language processing capabilities within the e-commerce industry.
%We conducted a study on the generalizability of LoRA merging, evaluating its effectiveness in integrating diverse information types and the performance of models on cross-task scenarios, facilitating insights into task transferability within the e-commerce domain. 

% Through rigorous experimentation and analysis, this paper provides the exploration of  the potential of LLMs in advancing natural language processing capabilities within the e-commerce landscape.

\end{abstract}

\ccsdesc[500]{Applied computing~E-commerce infrastructure}

\ccsdesc[500]{Computing methodologies~Natural language generation}

\ccsdesc[300]{Computing methodologies~Information extraction}

%%
%% Keywords. The author(s) should pick words that accurately describe
%% the work being presented. Separate the keywords with commas.
% TODO: Add keywords
\keywords{Large Language Models, LLM, E-Commerce, Text Generation}
%% A "teaser" image appears between the author and affiliation
%% information and the body of the document, and typically spans the
%% page.

\received{16 August 2024}
% TODO Change these dates if accepted
\received[revised]{30 August 2024}
\received[accepted]{25 October 2024}

%%
%% This command processes the author and affiliation and title
%% information and builds the first part of the formatted document.
\maketitle

\section{Introduction}
% \textcolor{red}{If we want to go fancy. 
% Selling point: we simulate common practice in e-commerce: training opensource state of the art model from scratch and compare with the feasibility of using LLM and see how if llm can bring in more gain in the evaluation training LLM does require a lot of resources; a common practice to efficiently train the model is to 
% }

Large Language Models (LLMs) have recently gained ubiquitous use in many domains\citep{singhal2022large,trautmann2022legal,malinka2023educational}.
In the e-commerce domain in particular, LLMs have the potential to facilitate the creation of product descriptions, summarize reviews, expand queries, and answer buyer and seller questions, among other potential use cases. 

We simulated a common practice in the e-commerce industry: adapting  open source state of the art models for domain specific tasks. 
We compared the feasibility of using an LLM and explored to what extent using an LLM for different e-commerce tasks leads to gains in the evaluation metrics. 
Training LLMs from scratch requires a significant amount of resources, but a common practice to efficiently train the model is to use parameter efficient methods like low rank adapters (LoRA) \citep{hu2021lora}. 
Important questions arise when attempting to adapt these models for specific tasks and domains. 
We attempt to answer in this work:  
How much training data is needed to adapt a model to a task? 
How much do LLMs outperform more traditional approaches?
What ways do different tasks interact with each other when doing mixed dataset training or merging LoRA weights trained on indivudal tasks?
There are various approaches for adapting LLMs for tasks in a specific domain. 
Specifically, we focus on LoRA supervised fine-tuning (SFT), multi-task training, zero-shot inference,  and LoRA merging. 

%Our contributions include: 1) Organizing and formatting four e-commerce datasets for LLM evaluation. 2) Conducting comprehensive experiments to compare the fine-tuning of a large language model with conventional industrial baselines, such as BERT and T5 using varying amounts of data for e-commerce tasks. 
%3) Comparing fine-tuning a smaller LLM with zero and few-shot inference with a larger LLM.
%4) Exploring mixed LoRA merging across different tasks and comparing it with more traditional mixed dataset training.

Our contributions are as follows: 1) Organizing and formatting four e-commerce datasets for the evaluation of large language models (LLMs). 2) Conducting comprehensive experiments to compare fine-tuning a large language model with conventional industrial baselines, such as BERT and T5, using varying amounts of data for e-commerce tasks; 
additionally, we examine the effectiveness of in-context learning with a highly competitive, very large LLM. 3) Exploring the use of mixed LoRA merging across different tasks and comparing this approach to traditional mixed dataset training.

Our findings indicate that for e-commerce-specific tasks, conventional methods, such as training smaller language models, can achieve performance that is comparable to or even surpasses that of general-purpose very large LLMs. These valuable insights provide  for the application of these models within the e-commerce industry.

\section{Background}
Large Language Models (LLMs) \citep{chowdhery2023palm, achiam2023gpt, anil2023palm, almazrouei2023falcon, touvron2023llama} have seen increasing attention in recent years as models that perform natural language generation have begun to be used for multiple tasks.
They differ from prior pre-trained language models (PLMs), such as BERT \citep{devlin-etal-2019-bert} or T5 \citep{raffel2020exploring}, in their amount of training data and number of parameters.

\subsection{Instruction Fine-tuning}
% With the recent wave of the large language models, application of the LLM into a specific domain is important to significantly expand the spectrum of the LLM into the industrial application. 
Instruction fine-tuning represented a pivotal advancement in the optimization of large language models (LLMs), such as GPT-4, for enhanced task-specific performance, especially in domain-specific applications
\citep{hu-etal-2023-llm,zhang2024scaling}.
This method involved the supplementary training of a pre-trained based model such as GPT \citep{achiam2023gpt}, Llama \citep{touvron2023llama}, or Falcon \citep{almazrouei2023falcon} on a task specific dataset consisting of prompts paired with their optimal responses. 
The objective was to refine the model's capacity to comprehend and execute instructions with increased accuracy and contextual relevance.
Instruction fine-tuning has emerged as an invaluable technique for augmenting the proficiency of LLMs across various specialized domains, ensuring their outputs align more closely with user expectations and requirements. 
% This approach underscored the evolving landscape of artificial intelligence, where tailored model training methodologies enhance both the versatility and applicability of LLMs.

% Fine-tuning LLM has been one of the most prevalent practice in the application of LLM, especially in domain-specific applications
% \citep{hu-etal-2023-llm,zhang2024scaling}.
% The publicly released base model trained on the general next token prediction task can be fine-tuned to learn domain-specific knowledge while still maintaining the basic ability of an LLM. 

\subsection{Low-Rank Adaptation Training}
Low-Rank Adaptation (LoRA) \citep{hu2021lora} is an innovative technique designed to fine-tune (LLMs) in a resource-efficient manner. 
This method addresses the challenge of adapting pre-trained models to specialized tasks without the extensive computational costs associated with traditional full-model fine-tuning. 
At the heart of LoRA is the strategic introduction of trainable low-rank matrices that target specific components of the LLM's architecture, 
namely the attention and feed-forward neural network layers inherent to the transformer model.
Specifically, it freezes the pre-trained layers of the LLM, and for each layer, it trains a rank-decomposition matrix and injects them into each layer of the pre-trained model to accomplish the LLM fine-tuning.  

LoRA involves the addition of low-rank matrices $\mathbf{A}$ and $\mathbf{B}$ to the existing weight matrices $\mathbf{W}$ of the model.
The model's original parameters are kept frozen during fine-tuning while $\mathbf{A}$ and $\mathbf{B}$ are updated. 
These matrices $\mathbf{A}$ and $\mathbf{B}$ are much smaller in size compared to $\mathbf{W}$, enabling significant reductions in the number of trainable parameters. 
The adaptation occurs through the equation. $\mathbf{W}' = \mathbf{W} + \mathbf{A}*\mathbf{B}^T$, where $\mathbf{W}'$ represents the adapted weights. This process selectively fine-tunes the model, allowing it to acquire new capabilities or improve performance on specific tasks with minimal adjustments to its pre-trained parameters. 
This selective updating is particularly beneficial for domain-specific applications, where only certain aspects of the model's knowledge need refinement. 

\subsection{Evaluation}
With the rise of text generation models that are seemingly capable of performing large numbers of tasks and able to answer many questions, a number of evaluation strategies have been proposed.
Evaluation leaderboards often consist of evaluation tasks like Hellaswag \citep{zellers-etal-2019-hellaswag}, MMLU \citep{hendryckstest2021}, and others \citep{eval-harness,hendryckstest2021,hendrycks2021ethics} which cover a broad range of multiple choice questions. 
These types of multiple choice question evaluations, where the answer is chosen based on the choice with the highest likelihood can be brittle as rankings can be sensitive to minute details \citep{alzahrani2024benchmarks}. 
Other evaluation benchmarks like GLUE \citep{wang-etal-2018-glue} consist of a bundle of different tasks with task specific evaluation metrics.
Another approach to evaluating LLM performance is the approach of LLM as a judge. 
Chatbot arena \citep{zheng2024judging} is an example of this style of evaluation and while it can correlate with human judgment, like other LLM applications it can be subject to hallucinations.
In this work, we focus on directly evaluating the tasks of interest with existing scoring practices.

\subsection{LLMs for E-commerce}
\citet{zhang2024scaling} find that fine-tuning LLM scaling factors appear to be very task dependent; however, there has been little published work examining fine-tuning focusing on the e-commerce domain. 
There has been some prior work investigating the use of LLMs on e-commerce tasks. 
ECOMGPT \citep{li2023ecomgpt} looked at framing e-commerce tasks as instruction fine tuning, but doesn't explore LoRA \citep{hu2021lora} or how different tasks enhance or interfere with each other or the \emph{amount} of data required for reasonable performance.

\section{E-Commerce Datasets}
There are a limited number of e-commerce datasets publicly available. 
Currently, there are few e-commerce benchmarks for evaluating LLMs on e-commerce tasks.  
We collected four datasets covering classification, sequence labeling, and product description  generation, and review summarization in order to evaluate the performance of LLMs in the e-commerce domain. 

\begin{table}[tbh]
\centering
\footnotesize
\begin{tabular}{@{}lrrr@{}}
\toprule
 Task & \multicolumn{1}{l}{Train} & \multicolumn{1}{l}{Dev} & \multicolumn{1}{l}{Test} \\ \midrule
ESCI Classification & 1,393,063   & -   & 425,762  \\
QueryNER: Query Segmentation                     & 7,841                     & 871                     & 933                      \\
Review Summarization                  & 25,203                    & 3,114                   & 3,166                    \\
Description Generation & 431,470 & - & 103,865 \\

% hetPQA & 1,834                     & 186                     & 2,289                    \\ 
\bottomrule
\end{tabular}
\caption{Sizes of dataset splits by input prompt and output pair for each task.
}
\label{tab:dataset-size}
\end{table}

\subsection{ESCI Multi-class Product Classification}
The Shopping Queries ESCI dataset \citep{reddy2023shopping} contains search queries, released with the aim of fostering research in the area of semantic matching of queries and products. 
The dataset contains three tasks: Query-Product Ranking, Multi-Class Product Classification,  Product Substitute Identification. 
We use the ESCI Multi-Class Product Classification task.
The task is to classify a query and product pair as an exact match (E), a substitute (S), a complementary product (C) or Irrelevant (I).
Because Query-Product Ranking and Product Substitute Identification involve more complexity and longer input strings, we do not include them for LLM evaluation in this work. 
%%% This next piece is covered later.
% We reformat the classification task with an input prompt and the text of the product to be classified. 
% The output string is just the classification label as a string, so the model only needs to generate a single word in order to make predictions.

\subsection{QueryNER}

% TODO: Add new citation for QueryNER
QueryNER \citep{palen-michel-etal-2024-queryner} is an e-commerce query segmentation dataset.
The task in QueryNER is not to extract aspects, but rather to segment the user’s query into meaningful chunks. 
QueryNER uses a subset of the Amazon Shopping Queries Dataset \citep{reddy2023shopping} as the underlying data.
QueryNER consists of an ontology of 
17 types.
The entity types are: core\_product\_type, product\_name, product\_number, modifier, creator, condition, unit of measurement (UoM), department, material, time, content, color, shape, quantity, occasion, origin, and price. 

Because QueryNER uses BIO encoding (\textit{B} for begin, \textit{I} for inside a span, \textit{O} for outside a span), we linearized the data in order to have an input prompt and output string. 
The formatting of the linearized data is a series of (token, label) pairs. 

% For scoring purposes, we use a regular expression to extract the predicted labels from the output. 
% Because we noticed the model sometimes being inconsistent with the output format, the regular expression also handles comma separated output without parentheses. 
% In the case of further formatting issues or when the model does not predict labels for all tokens, it is assumed the model failed to generate a valid label sequence and that particular example is assigned all Os. 
% \citet{palen-michel-etal-2021-seqscore} highlighted issues with NER scoring.
% For scoring procedure clarity, once the model output has been extracted from the linearized form, we use seqeval \citep{seqeval} for evaluation using the setting is equivalent with conll. 

\subsection{Review Summarization}
AMASUM \citep{brazinskas-etal-2021-learning} is a dataset for summarizing product reviews.
The product reviews are in English and come from bestreviews.com, cnet.com, pmag.co.uk, runrepeat.com, which mainly consist of electronics and running shoes reviews. 
The dataset contains a list of product reviews and a summary with the ``verdict" on the product and also lists of pros and cons. 
The original paper focuses on selecting useful reviews in order to summarize. 

The goal of our evaluation of LLM performance is not to assess its review selection capability, so we select a small number of reviews to summarize. 
We select only 4 reviews to be used to generate the verdict summary. 
Since the dataset has a field with helpful votes where users voted that the review was helpful, we take the top four reviews with the most helpful votes as our selection process.

%%% We never used pros and cons summaries so skip.
% We select reviews for pros and cons summarization using the same process but filter by the rating so that reviews for cons have a rating below 3.0 and for pros above 3.0. 

\subsection{Product Description Generation}
There appears to be a lack of standard benchmark datasets in English for product description generation despite a decent amount of prior work. 
\citet{koto-etal-2022-pretrained} stated they were not able to release the dataset due to copyright issues. 
\citet{chan-etal-2019-stick} and \citet{zhang2019automatic} collected data from Taobao\footnote{\url{https://www.taobao.com/}}.
\citet{zhang2019automatic} also stated that there was no other standard dataset for product description generation. 
\citet{wang-etal-2017-statistical} created their dataset from attribute values and descriptions from Amazon but only in the category "Computers and Tablets".  

Without a clear prior benchmark for this task, we create a simple product description task from the ESCI Shopping Queries Dataset \citep{reddy2023shopping}.
We assemble an input consisting of a product title, brand, color, and bullet points. 
The bullet points in the original dataset are aspect-value pair information about a product or short snippets about the product. 
The expected output is the product description. 
We filter out items where there is no title, no description, no bullet points, or items where the description is an exact match of the title or bullet points.

% \subsection{Product Question Answering}
% For the question answering task we used hetPQA \citep{shen-etal-2022-product} since it provided a question, answer, and also candidate evidence to support the answer.
% \citet{shen-etal-2022-product} originally created the dataset as two separate tasks. 
% The first task was evidence ranking where the goal is just to determine if a piece of evidence, for example a product attribute or review, was relevant to the question. 
% The second task was answer generation. 
% Answer generation took whatever the relevant piece of evidence was and generated a natural language version of it that answers the question. 

% For framing these tasks for an LLM, we chose to frame product question answering as an end-to-end task. 
% The LLM is given the question and all the evidence available as input and must generate an answer as output.

% In order to create these end-to-end examples, we take each question for the answer generation task and check if there is evidence for the question in the evidence ranking task using the question id. 
% If there is evidence for a question, we create an instance with the question, evidence, and answer.

\subsection{Task Alignment \& Prompt Design}
\label{sec:taskalignment}
To enable instruction fine-tuning, each individual dataset was required to be aligned to a sequence to sequence style task in order to be used with an LLM.
Review summarization and product description generation already were easily treated as sequence to sequence tasks. 
However it was less obvious how to best treat classification and sequence labeling tasks. 
We treated the classification label as text to be generated. 
For sequence labeling, the model was expected to output tuples of each token along with its label. 

Prompts were designed to provide the model with enough context to accomplish its task. 
We created task specific templates for each of the tasks.
Each prompt asked the model to act as an e-commerce expert to provide context. 
The prompt then gave a brief description of the task and along with the training example.
Examples of prompts are shown in Appendix \ref{sec:appendix}.

\section{Experiments}

\subsection{Baselines}
We chose to use the most common and competitive baselines for each e-commerce task. 
For classification tasks, ESCI Task 2 and QueryNER, we chose to use BERT (BERT-base) \citep{devlin-etal-2019-bert} as the baseline model with a  learning rate of 3e-5 following \citet{devlin-etal-2019-bert}.
The training of BERT followed the conventional formulation of the sequence classification problem for the ESCI task and token classification for the QueryNER task. 

For generative tasks, review summarization, and product description generation, we chose to use T5 (T5-base) \citep{raffel2020exploring} as a baseline model. 
T5 and BERT are fine-tuned with default parameters released by Hugging face \citep{wolf-etal-2020-transformers}.
Because in-context learning has been shown to be effective \citep{dong2024surveyincontextlearning}, we use we include Mixtral 8 x 22b \citep{jiang2024mixtralexperts} as a theoretical state of the art zero-shot and few-shot baseline.

% For our purposes, inference with a larger model was doable, but SFT on a model of this size was less practical. 

\subsection{Mix Tasks Training}
Mixing tasks (datasets) for fine-tuning large language models (LLMs) can enhance the model's performance, generalization ability, and adaptability to various tasks, which closely mirrors the industrial application. 
The trained model was not required to accomplish one task but rather several domain-specific tasks such as query named entity recognition (NER), text summarization, description generation, and classification. Fine-tuning LLMs with mixed and diverse training datasets could help improve performance on each task.

\subsection{Mix LoRA Merging}
Furthermore, thanks to the nature of the LoRA framework, mixed tasks (datasets) training could also be achieved by merging LoRA weights independently trained on different tasks. Specifically, assume there were \(n\) tasks with a specific task number \(i \in \{0, \ldots, n-1\}\). LoRA weights trained independently from each task could be defined as \(\mathbf{A}_{i}\) and \(\mathbf{B}_{i}\). The adaptation equation for mixed LoRA merge was \(\mathbf{W}' = \mathbf{W} + \frac{1}{n} \sum_{i=0}^{n-1} \mathbf{A}_i \mathbf{B}_i^T\). 
Mixing LoRA merging provided additional flexibility since additional tasks could be added later instead of retraining whenever a new task was added.
Additionally, some types of tasks may benefit each other, while others may lower another tasks performance when merged. 
Mixing LoRA merge enabled more efficient experimentation with different combinations of tasks compared with directly training with the mixed data set.

\subsection{Implementation Details}
The foundation model used in all LoRA fine-tuning experiments was Llama-2-7b as its moderate size and considerable performance on various tasks \citep{touvron2023llama}. The supervised fine-tune was conducted with $3$ epochs for each dataset, as we do not see performance gain while we trained it longer. Since all four tasks for the LLM were formulated as a text generation task (see Section \ref{sec:taskalignment}), we followed the most common hyperparameters to finetune the LLM under the text generation setup. Specifically, the model was loaded with $8$ bit, the max length of input was set to $1500$, and the precision of the parameter was set to be \textit{bf16}. During the supervised fine-tune, we adopted the LoRA \citep{hu2021lora} to conduct efficient model training. The LoRA $\alpha$ was set to be $16$, with dropout rate be $0.05$. The initial learning rate was
chosen to be $3e-5$ with a cosine learning rate scheduler. To further improve the training efficiency, we adopted paged\_adamw\_8bit as optimizer. The LLM model training were conducted on two NVIDIA A100 80GB GPUs. 

To optimize computational time, we randomly sampled subsets from each dataset for our experiments. Specifically, for the ESCI dataset, we used samples of 5k, 10k and 50k as the training set and sampled 1k from the original test set to serve as the test set. For the QueryNER dataset, we selected 1k, 5k, and 8k (full 7,841 data samples) samples for training and used the entire original test set as the test set. In the case of the description generation dataset, we randomly chose 1k, 5k, 10k, and 25k samples for training, with 10k samples from the original test set used as the test set. Finally, for the Review Summarization dataset, we sampled 1k, 5k, 10k, and 25k (the entire 25,203 dataset) as the training set and utilized the complete original test set as the test set. All experiments were conducted using the same aforementioned hyperparameters.

\section{Results}
\subsection{Metrics on different tasks}
Since the datasets we explored contained both classification and generative tasks, we use the task-specific metric to evaluate the performance of the model performance on various datasets. For classification tasks, we use F1 score as evaluation metrics and report both micro-average and macro-average results, while for the generative task, we use the Rouge-1 F1 (Rouge-1) and Rouge-L F1 (Rouge-L) \citep{lin2004automatic} to evaluate the performance. 

\subsection{Evaluating LLM on Classification Tasks}
To ensure accurate mapping of the generated text to the actual class label in the classification task, we implemented a simple but strict evaluation approach. 
For ESCI classification, the LLM output was required to match the corresponding label exactly. 
Any deviation from the exact label resulted in the classification being considered incorrect. For QueryNER, labels were extracted using a regular expression expecting a list of tuples of tokens with BIO tags.
If the LLM output deviated from the expected output, the labels were considered all \textit{O}s (no entities identified). 
Because we noticed the model sometimes being inconsistent with the output format, the regular expression also handles comma separated output without parentheses. 
In the case of further formatting issues or when the model does not predict labels for all tokens, it is assumed the model failed to generate a valid label sequence and that particular example is assigned all \textit{O}s. 
\citet{palen-michel-etal-2021-seqscore} highlighted issues with NER scoring.
For scoring procedure clarity, once the model output has been extracted from the linearized form, we use seqeval \citep{seqeval} for evaluation using the setting which is equivalent with conlleval. 

\subsection{SFT LLMs vs Baseline PLMs}
We performed SFT training of Llama2-7b on each dataset with different portions of the data to explore the impact of training data size for fine-tuning an LLM on each task.

\subsubsection{Classification tasks}
% \input{tab/tab_class}
%\input{tab/tab_esci_results}
%\input{tab/tab_ner}
% Please add the following required packages to your document preamble:
% \usepackage{booktabs}
% Please add the following required packages to your document preamble:
% \usepackage{booktabs}
\begin{table}[tbh]
\footnotesize
\begin{tabular}{@{}r|cc|c|cc@{}}
\toprule
               & \multicolumn{2}{c|}{ESCI Classification} &     & \multicolumn{2}{c}{QueryNER}    \\ \midrule
               & Micro F1            & Macro F1           &     & Micro F1       & Macro F1       \\ \midrule
BERT @5k       & 0.348               & 0.181              & @1k & 0.539          & 0.390           \\
LLM SFT @5k    & 0.355               & 0.244              & @1k & 0.280           & 0.156          \\ \midrule
BERT @10k      & 0.628               & 0.294              & @5k & 0.580           & 0.508          \\
LLM SFT @10k   & 0.397               & 0.213              & @5k & 0.553          & 0.398          \\ \midrule
BERT @50k      & \textbf{0.629}      & \textbf{0.368}     & @8k & 0.603          & 0.569          \\
LLM SFT @50k   & 0.628               & 0.294              & @8k & \textbf{0.626} & \textbf{0.579} \\ \midrule
Mixtral 0-shot & 0.571               & 0.199              &     & 0.145          & 0.063          \\
Mixtral 3-shot & 0.537               & 0.009              &     & 0.484          & 0.336          \\ \bottomrule
\end{tabular}
\caption{Comparison of Llama2-7b SFT, finetune BERT and in context learning using Mixtral 8 x 22b model on ESCI task 2 dataset and QueryNER dataset.}
\label{tab:ESCI-queryNER}
\end{table}

Table \ref{tab:ESCI-queryNER} shows the performance comparison on classification tasks on ESCI Classification and QueryNER dataset among Llama2-7b Supervised fine-tuning, BERT model fine-tuning and in context learning using Mixtral 8 x 22b in zero and few-shot setup.  
Note that, instead of generating the distribution like BERT, the task for the LLM is to generate the classification result in text. As the number of the training samples increased the performance of the model generally increased. However, there was a clear performance boost of LLM as the number of training samples increased (from 10k to 50k on ESCI task 2 dataset and from 1k to 5k on Query NER dataset). In general, the LLM and BERT performed comparable in these classification tasks when given sufficient training data.

In domain-specific tasks such as ESCI classification and Query NER, the application of in-context learning with very large language models like the Mixtral 8x22b often does not meet the performance benchmarks achieved through fine-tuning. Despite the introduction of extensive context, these models frequently struggle to deliver the level of accuracy required for industrial applications. This observation underscores a critical limitation: while LLMs are versatile and powerful, they may not be inherently optimized for tasks that demand high precision within a specialized domain.

In contrast, fine-tuning enables models to be specifically tailored to the intricacies of the task at hand, allowing for a deeper understanding of domain-specific patterns and nuances. As a result, training smaller, task-specific models such as BERT, particularly those employing a softmax classification layer, often emerges as a more effective strategy. These models not only demonstrate superior performance but also offer advantages in computational efficiency, making them more suitable for deployment in resource-constrained industrial environments where both accuracy and efficiency are paramount.

\subsubsection{Generation task}

%\input{tab/tab_summ_gen}
% Please add the following required packages to your document preamble:
% \usepackage{booktabs}
% Please add the following required packages to your document preamble:
% \usepackage{booktabs}
\begin{table}[tbh]
\centering
\footnotesize
\begin{tabular}{@{}r|cc|cc@{}}
\toprule
                 & \multicolumn{2}{c|}{Review Summarization} & \multicolumn{2}{c}{Desc. Generation} \\ \midrule
                 & Rouge-1             & Rouge-L             & Rouge-1           & Rouge-L          \\ \midrule
T5 @1k           & 0.155               & 0.137               & 0.239             & 0.216            \\
LLM SFT @1k      & 0.158               & 0.147               & 0.262             & 0.244            \\\midrule
T5 @5k           & 0.162               & 0.147               & 0.241             & 0.223            \\
LLM SFT @5k      & 0.182               & 0.161               & 0.249             & 0.232            \\\midrule
T5 @10k          & 0.163               & 0.150                & 0.238             & 0.221            \\
LLM SFT @10k     & 0.186               & 0.162               & 0.258             & 0.241            \\\midrule
T5 @25k          & 0.169               & 0.158               & 0.232             & 0.215            \\
LLM SFT @25k     & \textbf{0.187}               & \textbf{0.165}               & 0.237             & 0.222            \\ \midrule
Mixtral 0-shot   & 0.099               & 0.090                & 0.248             & 0.229            \\
Mixtral 3-shot & 0.144               & 0.131               & \textbf{0.274}             & \textbf{0.254}            \\ \bottomrule
\end{tabular}
\caption{Comparison of Llama2-7b SFT with fine-tune T5 and zero-shot Mixtral 8 x 22b on Review Summarization dataset and Product
Description Generation dataset.}
\label{tab:sum}
\end{table}

Table \ref{tab:sum} shows the performance comparison of text generation tasks on the review summarization and description generation datasets. Similar to the classification tasks, there was a significant increase in model performance as more data samples were used for training. Notably, the  LLM consistently outperformed the conventional T5 model across both datasets. This superior performance can be attributed to the LLM's larger model capacity and enhanced quality of pre-training.

Fine-tuned models (Llama2-7b and T5) outperformed the zero-shot capabilities of the much larger Mixtral 8x22B model in review summarization, while for description generation, the performance is comparable. Despite the Mixtral model's strong standing on LLM leaderboards like \citep{open-llm-leaderboard-v2}, which suggests competitive summarization abilities, the observed performance gap between zero-shot and few-shot scenarios highlights a key limitation: without in-context guidance, the model struggles to achieve sufficient capability on domain-specific tasks (Review Summarization). However, when in-context information is provided, the model demonstrated significantly improved outcomes. In review summarization, to achieve even higher levels of performance, task-specific training becomes crucial. Notably, even with smaller model architectures, fine-tuning can yield superior results (using Llama2-7b). 

In contrast, the description generation task is more aligned with general-purpose text generation, where the model's ability to understand and leverage general knowledge is the primary factor in determining performance. Consequently, in this task, larger models like Mixtral, equipped with in-context guidance, could achieve top-tier performance, even surpassing fine-tuned smaller models.

%Few-shot appears to perform better for description generation than fine tuning, while for the summarization task, fine-tuning appears to perform better.
\begin{table}[tbh]
\tiny
\centering
\begin{tabular}{@{}
>{\columncolor[HTML]{FFFFFF}}l |
>{\columncolor[HTML]{FFFFFF}}c 
>{\columncolor[HTML]{FFFFFF}}c |
>{\columncolor[HTML]{FFFFFF}}c 
>{\columncolor[HTML]{FFFFFF}}c |
>{\columncolor[HTML]{FFFFFF}}c 
>{\columncolor[HTML]{FFFFFF}}c @{}}
\toprule
& \multicolumn{2}{c|}{\cellcolor[HTML]{FFFFFF}QueryNER}  & \multicolumn{2}{c|}{\cellcolor[HTML]{FFFFFF}Review Summ.} & \multicolumn{2}{c}{\cellcolor[HTML]{FFFFFF}Desc. Generation} \\ \midrule
& Micro F1                 & Marco F1               & Rouge-1                  & Rouge-L                 & Rouge-1                       & Rouge-L                      \\
\midrule
QueryNER @ 5k                   & 0.553                     & 0.398                   & -                        & -                       & -                             & -                            \\
+ Summ. LoRA @ 5k            & 0.002                    & 0.232                  & 0.192                    & 0.164                   & -                             & -                            \\
+ Desc. Generation @ 5k      & 0.018                    & 0.344                  & -                        & -                       & 0.251                         & 0.233                        \\ \midrule
                           & \multicolumn{2}{c|}{\cellcolor[HTML]{FFFFFF}ESCI} & \multicolumn{2}{c|}{\cellcolor[HTML]{FFFFFF}Review Summ.} & \multicolumn{2}{c}{\cellcolor[HTML]{FFFFFF}Desc. Generation} \\ \midrule
ESCI @ 5k                  & 0.355                    & 0.244                  & -                        & -                       & -                             & -                            \\
+ Summ. LoRA @ 5k            & 0.145                    & 0.174                  & 0.184                    & 0.156                   & -                             & -                            \\
+ Desc. Generation LoRA @ 5k & 0.137                    & 0.299                  & -                        & -                       & 0.239                         & 0.221                        \\ \bottomrule
\end{tabular}
\caption{Results of merging LoRA for different pairs of tasks.}
\label{tab:merge}
\end{table}
\label{sec:lora_merge}
\subsection{LoRA Merge}

We experimented with merging different pairs of LoRA weights for each pair of tasks. 
For this experiment, we used the LoRA weights from the 5k training set size.
To merge the LoRA weights we took the average of the two. 
The results of merging LoRA weights, shown in Table \ref{tab:merge}, demonstrated that when weights trained on a task requiring a more rigid structure in the output like ESCI classification or QueryNER, the performance on those tasks suffers. 
However, the performance of description generation and review summarization remained comparable with the performance from independent training with the same number of examples.

Upon reviewing model output, we found that at least a portion of this degradation in performance on the tasks requiring a more strict output format was attributable to the output formatting requirements. 
However, some of this apparent degradation may not truly be quite as bad as it appears. 

In Section \ref{sec:improper-format} of the appendix, we show an example of model output for the QueryNER task was shown where the model did in fact output BIO labels some of which are correct labels, but with formatting unrecognized by the scoring script.
% If using low rank training our hypothesis is that.. fine tuning on one task Low rank space is fine, but training with mix task training doesn't improve the performance. Maybe too much information to capture it in the low rank space.  Could cite the LoRA paper here maybe?
%Merge with classification tasks classification gonna die. merge with text generation tasks and it does ok.

\subsection{Comparison with Mix Tasks Training}
% Please add the following required packages to your document preamble:
% \usepackage{booktabs}
% \usepackage[table,xcdraw]{xcolor}
% Beamer presentation requires \usepackage{colortbl} instead of \usepackage[table,xcdraw]{xcolor}
\begin{table}[tbh]
\footnotesize
\centering
\begin{tabular}{@{}c|c|c|c|c@{}}
\toprule
& \begin{tabular}[c]{@{}c@{}}ESCI Class.\end{tabular} & \begin{tabular}[c]{@{}c@{}}QueryNER \end{tabular} & \begin{tabular}[c]{@{}c@{}}Review Sum.\\ \end{tabular} & \begin{tabular}[c]{@{}c@{}}Desc. Gen.\end{tabular} \\ 
\midrule
     & Micro  F1                                                         & Micro  F1                                                        & Rouge-L                                                        & Rouge-L  \\
\midrule
Indep. train       & 0.355                                                            & \textbf{0.553}                                                          & 0.161                                                          & \textbf{0.232}                                                               \\

Mix train     & 0.342                                                            & 0.455                                                          & \textbf{0.163}                                                           & 0.218                                                                \\
Mix LoRA & 0.251                                                            & 0.001                                                          & 0.145                                                           & 0.174                                                                \\
\bottomrule
\end{tabular}
\caption{Comparison of independent task training with BERT / T5, mixed task training with LoRA Llama2 7b, and mixed LoRA merging for Llama2 at 5k examples for each task}
\label{tab:mix}
\end{table}
We merged the LoRA (Low-Rank Adaptation) weights from all four tasks for inference and performance analysis on each respective test set. The weights were chosen based on training with 5k samples from each dataset to ensure balanced information contribution. To establish a fair comparison between training with mixed data and mixing through LoRA weight merging, we trained the foundation model (Llama-2 7B) on 5k samples from each dataset, resulting in a total of 20k training samples, using the same hyperparameters as in the previous experiments.

The results, detailed in Table \ref{tab:mix}, indicated that mixed dataset training resulted in lower performance compared to training each task independently. This performance decrease was attributed to the limited capacity of the LoRA adapter and the distinct nature of each task, which deteriorated the model's ability to consistently produce outputs in the correct format especially in the classification tasks. Furthermore, the LoRA weight merging approach generally showed inferior performance compared to both mixed dataset training and independent task training. In mixed LoRA merging, classification tasks particularly suffered, with output format issues noted (see Section \ref{sec:lora_merge}). 
However, for text generation tasks, the performance remained competitive.
% Despite being a larger model, zero-shot with mixtral was only the strongest performing approach for the classification task. 
% However, even mixtral's zero-shot performance on this task is outperformed when the number of examples increases to 50k for both LoRA tuning of Llama 2 and fine-tuned BERT as shown in Table 2. 

% 0.25 * ESCI task 2 + 0.25 * QueryNER + 0.25 * Review Summariation + 0.25 * Product Description Generation

% Task merge each other. 

% 0.5 * ESCI task 2 + 0.5 * QueryNER

% 0.5 * ESCI task 2 + 0.5 * Review Summariation

% 0.5 * ESCI task 2 + 0.5 *Product Description

% 0.5 * QueryNER + 0.5 * Review Summariation

% 0.5 * QueryNER + 0.5 * Product Description 

% 0.5 * Review Summariation + 0.5 * Product Description

\section{Conclusion}
In this paper, we explored the application potential of LLMs in addressing common e-commerce tasks, benchmarking against conventional industrial models like BERT and T5. We collected four e-commerce datasets, containing classification and text generation tasks, and adapted both task types to a text generation framework for LLMs. 
Our findings reveal that: (1) Zero-shot with a larger LLM is not a clear win and that smaller models fine-tuned on a specific task can have better task specific performance. 
(2) LLMs required a certain volume of training data to reliably produce the correct formats in classification tasks,  yet they achieve competitive performance against the conventional BERT baseline and surpass the T5 baseline in text generation tasks. 
(3) Mix task training appears to perform slightly worse but comparable to independent training.
(4) While LoRA merging for classification tasks did not consistently maintain output format, it demonstrated that merging in text generation tasks could still deliver competitive performance on individual tasks.
(5) Few shot performance seems to be better than zero shot or fine tuning with very small amounts of data, but in all tasks except description generation, fine-tuning a model with enough data still performed better. 
Overall, we demonstrated that for specific tasks in the e-commerce domain, out-of-the box zero-shot LLM inference like our use of mixtral 8 x 22b does not outperform fine-tuning approaches on target tasks and there are opportunities for further exploration of adapting LLMs to the e-commerce domain.

%%
%% The acknowledgments section is defined using the "acks" environment
%% (and NOT an unnumbered section). This ensures the proper
%% identification of the section in the article metadata, and the
%% consistent spelling of the heading.
% \begin{acks}
% To Robert, for the bagels and explaining CMYK and color spaces.
% \end{acks}

%%
%% The next two lines define the bibliography style to be used, and
%% the bibliography file.
\bibliographystyle{ACM-Reference-Format}
\bibliography{anthology,colm2024_conference}
\clearpage
%%
%% If your work has an appendix, this is the place to put it.
\appendix
\section*{Appendix}
\label{sec:appendix}
\section{Example samples from each Dataset}
% Please add the following required packages to your document preamble:
% \usepackage{booktabs}
\begin{table*}[tbh]
\scriptsize
\begin{tabular}{@{}p{2cm} p{7cm} p{6cm}@{}}
\toprule
Task                & Prompt Input                  & Response                                                                                                   \\ \midrule
\makecell[lt]{ESCI \\Classification} &  \makecell[lt]{Query: revent 80 cfm\\Product: \\product\_title: Panasonic FV-20VQ3 \\ WhisperCeiling 190 CFM Ceiling Mounted Fan \\product\_brand: Panasonic \\ product\_color: White\\product\_bullet\_point:} 
\begin{itemize} 
\item WhisperCeiling fans feature a totally enclosed condenser motor and a double-tapered, dolphin-shaped bladed blower wheel to quietly move air
\item Designed to give you continuous, trouble-free operation for many years thanks in part to its high-quality components and permanently lubricated motors which wear at a slower pace
% \item Detachable adaptors, firmly secured duct ends, adjustable mounting brackets (up to 26-in), fan/motor units that detach easily from the housing and uncomplicated wiring all lend themselves to user-friendly installation
\item This Panasonic fan has a built-in damper to prevent backdraft, which helps to prevent outside air from coming through the fan
\item 0.35 amp  
\end{itemize}

 & Irrelevant                                                                                                      \\ \midrule
QueryNER            & yarn swift carrying case & (yarn, B-modifier), (swift, I-modifier), (carrying, B-core\_product\_type), (case, I-core\_product\_type) \\ \midrule
% TODO CHESTER: Fix this with a new example
\makecell[lt]{Description \\Generation} &
  \makecell[l]{Title: Delta Electronics RAD80L\\ BreezRadiance 80 CFM Heater/Fan/Light\\ Combo White (Renewed) \\  
  Brand: DELTA ELECTRONICS \\(AMERICAS) LTD. \\
  Color: White \\Bullet points: }
  
  \begin{itemize}
      \item Quiet operation at 1.5 sones
      \item Built-in thermostat regulates temperature. Energy efficiency at 7.6 CFM/Watt
      \item Precision engineered with DC brushless motor for extended reliability, this fan will outlast many household appliances
      \item Galvanized steel construction resists corrosion
       \item     Duct: Detachable 4-inch Plastic Duct Adapter 
  \end{itemize} &
  This pre-owned or refurbished product has been professionally inspected and tested to work and look like new. How a product becomes part of Amazon Renewed, your destination for pre-owned, refurbished products: A customer buys a new product and returns it or trades it in for a newer or different model. That product is inspected and tested to work and look like new by Amazon-qualified suppliers. Then, the product is sold as an Amazon Renewed product on Amazon. If not satisfied with the purchase, renewed products are eligible for replacement or refund under the Amazon Renewed Guarantee. \\ \midrule
\makecell[lt]{Review \\Summarization} &
  I expected a lot more from this product. When something says it will ""thicken"" or give lots of body \& fullness, most products deliver. This one does not. I have very fine thin hair \& this barely gave me an ounce of body. Very disappointed \& not worth the \$. =(I purchased at a salon before. This bottle from Amazon, was not as thick. Like it had been watered down. I wouldn't buy again. I was buying this at a local spa but they stopped stocking it so I did what every desperate person does-went to Amazon to see if they carried it. I've been using this product for years and it solves my thinning-hair problem perfectly with no stiffness and no nasty smell.My favorite blow dry gel. Soft and natural looking. Smells fresh and nice. &
  A good budget spray gel option for those who want a quality hold without the higher price tag. \\

\bottomrule
\end{tabular}
\caption{Instructions for each dataset to guide the SFT.}
\label{tab:examples}
\end{table*}
In Table \ref{tab:examples}, an example from each dataset is shown. 

\section{Prompt}
% Please add the following required packages to your document preamble:
% \usepackage{booktabs}
\begin{table*}[tbh!]
\scriptsize
\begin{tabular}{@{}p{4cm}|p{12cm}@{}}
\toprule
Task                & Template                                                                                    \\ \midrule
ESCI Classification & Act as an e-commerce expert. Given a query and a related product retrieved for this query, the goal of this task is to classify the product as being an Exact, Substitute, Complement, or Irrelevant match for the query. Query: \{The input query string\} Product: \{The product title\}  ... Please answer with label ``Exact", ``Substitute", ``Complement" or ``Irrelevant" only.                                                                 \\ \midrule
QueryNER &
Act as an e-commerce expert. Identify the entities in the following query. Use BIO tagging format as if this is a named entity recognition or chunking task. The format should be tuples of the token followed by the tag. For example: (air, B-modifier), (tight, I-modifier), (containers, B-core\_product\_type), (for, B-modifier), (food, I-modifier). 
  
  The entity types are: 
  UoM, color, condition, content, core\_product\_type, creator, department, material, modifier, occasion, origin, price, product\_name, product\_number, quantity, shape, time
  Query: \{The input query string\}\\
  \midrule
Descr. Generation   & \makecell[{{p{10cm}}}]{Act as an e-commerce expert. Given the following information about a product, generate a description for the product.\\
Title: \{Product Title\}\\
Brand: \{Brand\}\\
Color: \{Color\}\\
Bullet points: \{bullets\}} \\ \midrule
Review Summarization       & Act as an e-commerce expert. Summarize the following product reviews.                       \\ 
% \midrule
% hetPQA              & ???     \\ 
\bottomrule
\end{tabular}
\caption{Prompt templates used with each e-commerce task.}
\label{tab:prompts}
\end{table*}
In Table \ref{tab:prompts}, prompt templates for each task are shown. 

\section{Improperly Formatted LLM Output}
\label{sec:improper-format}

An example of model output for the QueryNER task is shown in Table \ref{tab:badexp} where the model did in fact output BIO labels some of which are correct labels, but with formatting unrecognized by the scoring script.

% Please add the following required packages to your document preamble:
% \usepackage{booktabs}
\begin{table*}[tbh]
\scriptsize
\begin{tabular}{@{}l|l@{}}
\toprule
Prompt     & \begin{tabular}[c]{@{}l@{}}Act as an e-commerce expert. Identify the entities in the following query. \textbackslash{}nUse BIO tagging format as if this iss\\ a named entity recognition or chunking task. \textbackslash{}nThe format should be tuples of the token followed by the tag. \textbackslash{}nFor example: (air,  \\ B-modifier), (tight, I-modifier), (containers, B-core\_product\_type), (for, B-modifier), (food, I-modifier). \textbackslash{}nThe entity types are::\\ \textbackslash{}nUoM, color, condition, content, core\_product\_type, creator, department, material, modifier, occasion, origin, price, product\_namm\\ e, product\_number, quantity, shape, time\textbackslash{}nQuery: sheets for queen size bed sale\end{tabular} \\ \midrule
Label      & \begin{tabular}[c]{@{}l@{}}(sheets, B-core\_product\_type), (for, B-modifier), (queen, B-UoM), (size, I-UoM), (bed, B-modifier), (sale, B-modifier)\end{tabular} \\ \midrule
Output & \begin{tabular}[c]{@{}l@{}}sheets (B-core\_product\_type), (for, B-modifier), (queen, I-modifier), (size, I-modifier), (bed, I-modifier), (sale, B-modifier)\end{tabular} \\ \bottomrule
\end{tabular}
\caption{Example of format discrepancy in the output from the LLM when merging the LoRA adaptor trained on the QueryNER and Description Generation datasets.}
\label{tab:badexp}
\end{table*}   

\end{document}